\documentclass[lettersize,journal]{IEEEtran}
\usepackage{amsmath,amsfonts}
\usepackage{algorithmic}
\usepackage{algorithm}
\usepackage{array}
\usepackage{textcomp}
\usepackage{stfloats}
\usepackage{url}
\usepackage{verbatim}
\usepackage{graphicx}
\usepackage{cite}
 \usepackage{epstopdf}
\usepackage{subcaption}
\usepackage{multirow}
\usepackage{multicol}
\usepackage{makecell}
\usepackage{wrapfig}
\usepackage{url}
\usepackage{color}
\usepackage{booktabs}
\usepackage{epigraph}
\usepackage{comment}
\usepackage{setspace}

\usepackage[hidelinks, breaklinks=true]{hyperref}
\usepackage{CJKutf8}
\begin{document}
\title{Signage-Aware Exploration in Open World using Venue Maps}

\author{Chang Chen, Liang Lu$^{\dagger}$, Lei Yang$^{\dagger}$, Yinqiang Zhang, Yizhou Chen, Ruixing Jia, Jia Pan$^{\dagger}$,~\IEEEmembership{Senior Member,~IEEE}
\thanks{Manuscript received: October 10, 2024; Revised: December 25, 2024; Accepted: January 22, 2025.}
\thanks{This paper was recommended for publication by Editor Markus Vincze upon evaluation of the Associate Editor and Reviewers’ comments.}
\thanks{$^\dagger$This work was partially supported by the Innovation and Technology Commission of the HKSAR government under the InnoHK initiative. (Corresponding authors: Liang Lu, Lei Yang, Jia Pan)}
\thanks{Chang Chen, Liang Lu, Lei Yang, Yizhou Chen, and Jia Pan are with Centre for Transformative Garment Production, Bldg 19W, Hong
Kong Science Park, Hong Kong, China, and also with the Department of
Computer Science, The University of Hong Kong, Pok Fu Lam, Hong
Kong, China. (changchen@link.cuhk.edu.cn, llu92@hku.hk, l.yang@transgp.hk, yz.chen@transgp.hk, jpan@cs.hku.hk)}%
\thanks{Yinqiang Zhang and Ruixing Jia are with the Department of Computer Science, The University of Hong Kong, Pok Fu Lam, Hong Kong, China. (zyq507@connect.hku.hk, ruixing@connect.hku.hk)}
\thanks{Digital Object Identifier (DOI): see top of this page.}
}

\markboth{IEEE ROBOTICS AND AUTOMATION LETTERS, PREPRINT VERSION, ACCEPTED April 2025}%
{Shell \MakeLowercase{\textit{et al.}}: A Sample Article Using IEEEtran.cls for IEEE Journals}


\maketitle
\begin{abstract}
Current exploration methods struggle to search for shops or restaurants in unknown open-world environments due to the lack of prior knowledge. Humans can leverage venue maps that offer valuable scene priors to aid exploration planning by correlating the signage in the scene with landmark names on the map. However, arbitrary shapes and styles of the texts on signage, along with multi-view inconsistencies, pose significant challenges for robots to recognize them accurately. Additionally, discrepancies between real-world environments and venue maps hinder the integration of text-level information into the planners. This paper introduces a novel signage-aware exploration system to address these challenges, enabling the robots to utilize venue maps effectively. We propose a signage understanding method that accurately detects and recognizes the texts on signage using a diffusion-based text instance retrieval method combined with a 2D-to-3D semantic fusion strategy. Furthermore, we design a venue map-guided exploration-exploitation planner that balances exploration in unknown regions using directional heuristics derived from venue maps and exploitation to get close and adjust orientation for better recognition. Experiments in large-scale shopping malls demonstrate our method's superior signage recognition performance and search efficiency, surpassing state-of-the-art text spotting methods and traditional exploration approaches. Project website: \href{https://sites.google.com/view/signage-aware-exploration}{\color{blue}sites.google.com/view/signage-aware-exploration}.

\end{abstract}


\section{Introduction}
\IEEEPARstart{H}{umans} can efficiently navigate and explore a mall to search for shops or restaurants using a 2D venue map provided by the mall or Google Maps, even if the venue map is non-metric, illustrating only relative relationships between the different landmarks. In contrast, robots struggle to search for landmarks (here we refer to shops or restaurants) in unknown environments due to a lack of scene priors. However, this capability is fundamental for various applications, such as delivery, tour guidance, and inspection. Recently, significant efforts have been made to improve robots by mimicking human behaviors, e.g.,
correlating the visual observations with venue maps for global localization \cite{wang2015lost, radwan2016you,howard2021lalaloc, sarlin2023orienternet, sarlin2024snap}, or by discovering non-metric heuristic derived from maps for kilometer-scale navigation \cite{shah2022viking, fan2023s2mat}. 
However, these approaches primarily focus on using geometric and semantic information from the venue maps but overlook the landmark names portrayed on the maps and the corresponding signage displaying the names in the scenes. 
Moreover, they often neglect online scene understanding and exploration. On the other hand, current exploration methods generally integrate geometric information~\cite{rrt, bircher2016receding} or object-level scene semantics~\cite{dang2018autonomous, chaplot2020object, papatheodorou2023finding, lu2024semantics, gu2023conceptgraphs, werby2024hierarchical}, but rarely incorporate text-level semantics, limiting their effective use of venue maps. While some efforts detect direction signs and arrows \cite{maye2010inferring, shaikh2013ocr, talbot2020robot}, and room names~\cite{case2011autonomous, mantelli2021semantic} as cues in public scenarios, they do not address the recognition of landmark signage. Observing that humans exploit these \textit{landmark names on the venue maps by matching them to the \textit{signage in the real-world scene},} we argue that a signage-aware exploration method is essential to improve a robot's ability to search for landmarks using 2D (non-metric) venue maps.

\begin{figure}[t]
    \centering
    \includegraphics[width=0.95\linewidth]{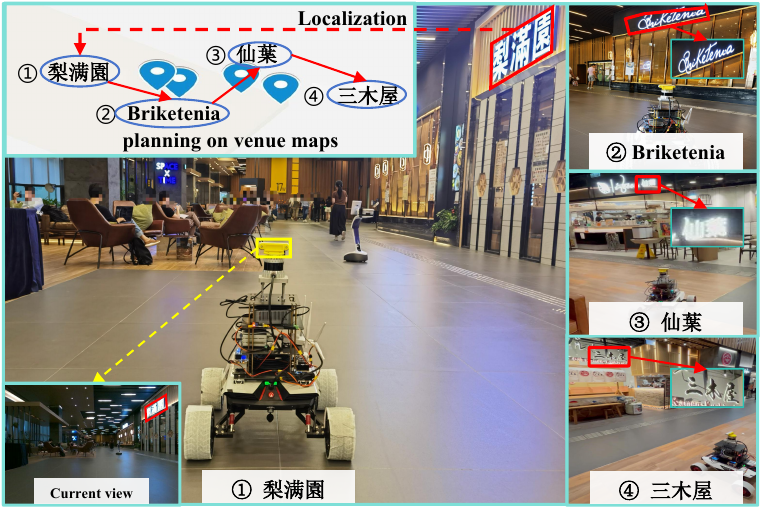}
    \caption{We propose to leverage the \textit{textual} information in a venue map to facilitate shop searching in unknown open-world environments. The robot localizes itself in the environment by recognizing and matching the texts on a sign to the venue map. Then the robot plans a direction to the next landmark `Briketenia'.
    }
    \vspace{-0.4cm}
    \label{fig:story}
\end{figure}

However, implementing such a system poses the following key challenges. 
First, recognizing the landmark names on the signage in an open-world environment is inherently difficult. Existing works in text recognition for visual localization \cite{wang2015lost, radwan2016you} or place recognition \cite{wang2015bridging, hong2019textplace, tang2022image, zimmerman2022robust, li2023textslam} often rely on optical character recognition (OCR) models trained on closed-set shapes and styles, which struggle to recognize signage effectively in open-world scenarios, where signage can exhibit diverse shapes, styles, and multi-view inconsistencies. 
Second, the accuracy of signage recognition also depends significantly on the robot's distance and orientation relative to the signage during movement, which requires an active perception policy to obtain clear observations. Third, using venue maps to guide exploration planning is challenging due to their inconsistent scales and distortions compared to real-world situations.

This paper presents a robotic exploration approach that leverages the signage in an unknown environment and the corresponding (non-metric) venue map to facilitate exploration and searching for landmarks. To address the first challenge, instead of relying on an OCR model trained to recognize texts in limited shapes and styles, we propose a human-like retrieval process based on the appearance of signage.
It detects the texts on the signage and matches them with a set of pre-generated signage images based on the landmark names extracted from the given venue map.
This allows a closed-set detector to adapt to open-world situations without fine-tuning. We also project multi-view 2D text regions into 3D space and fuse the corresponding features to enhance recognition performance. 
Furthermore, we design a venue map-guided exploration-exploitation planner that enables the robot to search for the signage corresponding to the landmarks using a directional heuristic. Once a sign candidate is detected, the system will exploit the known space to approach the sign and adjust the robot's view to faithfully recognize the sign. By balancing exploration and exploitation, we achieve both high signage coverage rates and search efficiency. Our main contributions to this work are summarized as follows:
\begin{itemize}
    \item A novel signage understanding method comprising a diffusion-based text instance retrieval and a 2D-to-3D semantic fusion strategy that can accurately detect and recognize the signage in the scene.
    \item A venue map-guided exploration-exploitation planner for efficiently covering the signage in the open world while grounding the signs in the map. 
    \item Evaluation experiments in real-world scenes have substantiated our method's efficacy and robustness.  
\end{itemize}

\section{Related works}

\subsection{Discovering the Potential of Venue Maps}
Venue maps, satellite maps, or OpenStreetMaps (OSM), are commonly encountered in our daily lives, offering valuable and readily accessible information for human and robot navigation. Existing works primarily study estimating the global pose $(x, y, \theta)$ on the 2D map's coordinate using visual observation over the latent space \cite{wang2015lost, radwan2016you, howard2021lalaloc, sarlin2023orienternet, sarlin2024snap}, such as SNAP \cite{sarlin2024snap}, which aligns the query image with a fuse of ground-level and overhead imagery over a learned 2D neural space, or inducing the potential heuristics from the map to achieve long-range point-goal navigation in open-world scenarios~\cite{shah2022viking, fan2023s2mat}. ViKiNG~\cite{shah2022viking} learns a heuristic function to predict whether future steps lie in the paths towards the goal, and S2MAT~\cite{fan2023s2mat} utilizes a user-specified GPS sequence to guide the navigation. 
As for leveraging the landmark names on the venue maps, \cite{wang2015lost} introduces the shop names into the global localization and formulates the problem as inference in a Markov random field, and \cite{radwan2016you} employs a particle filter into text detection for pose estimation. However, correlating shop names with the venue maps to facilitate exploration planning has not been studied.

\begin{figure}[t]
    \centering
    \includegraphics[width=\linewidth, height=0.45\linewidth]{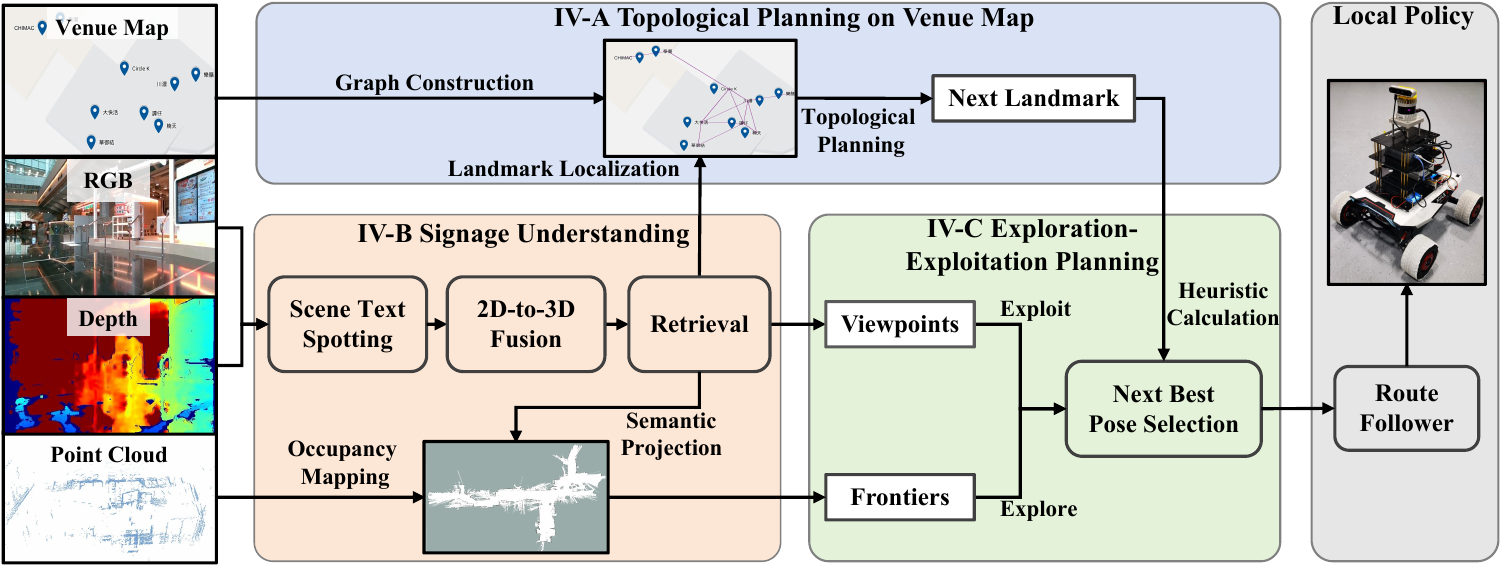}
    \vspace{-0.4cm}
    \caption{Overall framework. Our method first constructs a topological graph on a given venue map (Sec. \ref{planning}). Then, given the RGB-D image, the proposed signage understanding method recognizes the texts on the signage and correlates them with the text set of the venue map (Sec. \ref{sec:mapping}). Once localized on the venue map, the next landmark goal is inferred to guide the selection of frontiers. Our system balances exploration and exploitation to improve both signage coverage rates and search efficiency during the process (Sec. \ref{sec:exploration}).
    }
    \vspace{-0.2cm}
    \label{fig:overview}
\end{figure}

\vspace{-0.3cm}
\subsection{Semantic-aware Exploration}
Semantic-aware exploration methods that focus on the semantics in the scene have arisen in recent years \cite{dang2018autonomous, chaplot2020object, papatheodorou2023finding, lu2024semantics}. 
They integrate scene understanding methods with traditional planning, seeking both mapping efficiency and semantic coverage. SARHP \cite{lu2024semantics} proposes an object-centric planner for optimizing the semantic gain of the exploration paths. 3D scene graph methods that focus on the objects and their topological relations to enable high-level reasoning for robotic navigation and manipulation have also been developed, such as ConceptGraphs \cite{gu2023conceptgraphs} and HOV-SG \cite{werby2024hierarchical}. They segment the object masks via a proposal network, such as SAM~\cite{kirillov2023segment}, and extract the language-aligned features of the cropped regions via CLIP~\cite{radford2021learning} to support open-vocabulary query, which assumes a sequence of RGB-D trajectory is pre-collected manually or by any exploration method. Different from the above works that concentrate on object-level semantics, some studies target utilizing the direction signs and arrows \cite{maye2010inferring, shaikh2013ocr, talbot2020robot} or room names \cite{case2011autonomous, mantelli2021semantic} as navigation cues. Maye et al. \cite{maye2010inferring} introduce Hierarchical Implicit Shape Models for robustly recognizing different kinds of signs with only unsupervised training. Our work focuses on incorporating landmark names on the signage into exploration planning.

\vspace{-0.3cm}
\subsection{Text-based Place Recognition and Scene Text Spotting}
Text features have garnered significant attention \cite{wang2015bridging, hong2019textplace, tang2022image, zimmerman2022robust} in visual place recognition (VPR) tasks, which employs text-based descriptors to facilitate matching between a query image and a database of scene images, due to their robustness to the appearance changes and distinctive features in different places. TextSLAM \cite{li2023textslam} develops a novel SLAM system tightly coupled with text features to build a 3D volumetric map and enhance pose estimation accuracy. On the other hand, scene text spotting (STS) task that detects and recognizes the texts in natural scenes have also drawn great attention. ESTextSpotter~\cite{huang2023estextspotter} achieves state-of-the-art performance by explicitly synergizing the text detection and recognition modules.

Despite using the OCR models such as PP-OCRv3 \cite{li2022ppocrv3} and STS models \cite{huang2023estextspotter}, the signage recognition results are often too noisy due to the partial observation and multi-view inconsistency issues. Wang et al. \cite{wang2015lost} adopt a bag-of-N-gram model to mitigate the former issue, which requires training a 10000-dimensional classifier to predict whether each composition of the alphabet with a maximum length of N is the substring of the recognized texts and is still infeasible for square-shaped character languages that contain strokes. To make a general solution, we conduct similarity matching between the observed signage and the text set of venue maps to improve the signage recognition performance without fine-tuning. This is similar to the scene text retrieval (STR) task but with different motivations. Prior works in STR convert the queried text to an image via font rendering and perform image-level similarity matching to localize and retrieve all the text instances from an image gallery \cite{zhang2020adaptive, wang2021scene, wen2023visual}. In contrast, we improve the signage recognition performance and use a text-diffusion model for text-to-image conversion.

\vspace{-0.2cm}
\section{Problem Formulation}
We aim to design a signage-aware exploration method that leverages the venue map for searching for all the landmarks in unknown large-scale human-populated environments, such as a shopping mall, thereby proving our method's capability of navigating to any destination quickly. We model the environment as a 3D space $V \in \mathbb{R}^3$, where the points $\mathbf{v}\in V$ are classified into occupied, free, and unknown.
The environment contains $N$ static landmarks (shops or restaurants), which have the corresponding signs displaying their landmark names$\mathcal{T}$.
The venue map $\textbf{M}$ (see Fig. \ref{fig:story}) portrays all the landmarks and a topological graph $\mathcal{G}$ is generated among them. 
Since geometric information or semantic layouts may not be readily available in the given venue map, we do not make use of this information to maximize the generality of the proposed method.
The planner takes as inputs a venue map $\textbf{M}$, an RGB image $o_t$, a depth image $D_t$, and LiDAR points $\mathcal{X}_t$ at each timestep $t$, while concentrating on the signage in the scene. It outputs the next best pose $c_{t+1} \in \mathcal{F} \cup \mathcal{V}$ among the sampled frontiers $f \in \mathcal{F} $ and viewpoints $v \in \mathcal{V}$. During the process, a metric map with signage $\mathcal{M}$ (or simply a signage map) is constructed online, where the recognized signs $\tau \in \mathcal{T}$ are grounded in the corresponding regions of $\mathcal{M}$. The overall framework is illustrated in Fig. \ref{fig:overview}.

\vspace{-0.2cm}
\section{Methodology}

\subsection{Topological Planning on Venue Maps}
\label{planning}
We first pre-build a topological graph $\mathcal{G}$ on the given venue map whose nodes are the landmark names detected by an OCR model and whose edges connect two nodes if their Euclidean distance on the venue map is lower than a threshold $\gamma_{map}$. For any pair of isolated subgraphs, we merge them with their nearest nodes. As such, we solve a travel salesman problem (TSP) to obtain a landmark route $g_{1:K}$. During online exploration, our method searches for the remaining landmarks sequentially to handle long-horizon planning \cite{chen2024scale}.

\begin{figure}[t]
    \centering
    \includegraphics[width=0.8\linewidth]{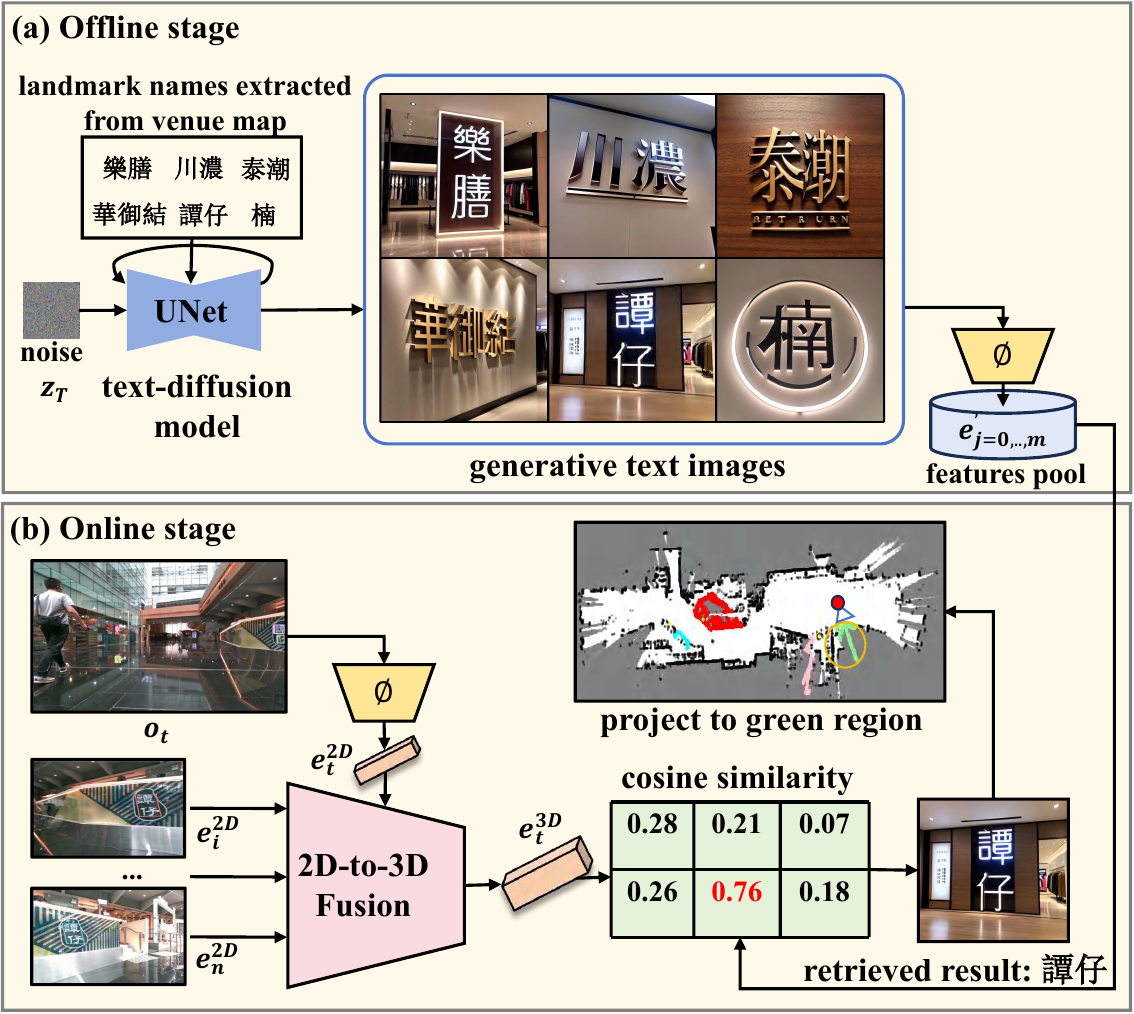}
    \caption{The pipeline of signage understanding. (a) In the offline stage, we use a text-diffusion model to render the landmark names on the generative images. (b) In the online stage, we project the detected text images to 3D space and fuse the features with those of multi-view images. Then we retrieve the most similar offline images compared to the detected images as the results, which are projected onto a signage map. All text features are extracted by a scene text spotter $\phi$.}
    \label{fig:scene-text-diffusion}
     \vspace{-0.2cm}
\end{figure}

\vspace{-0.2cm}
\subsection{Signage Understanding}
\label{sec:mapping}

We online detect and recognize the signage using the proposed signage understanding method, which enables the robot to globally localize itself on the venue maps. Meanwhile, we construct a signage map $\mathcal{M}$ for downstream querying and planning, which stores the recognized landmark names $\tau$ on the signage and the corresponding located regions. In practice, we assign the clustered contours of the online map that are nearest to the signage as its corresponding region.

\textbf{Diffusion-based Text Instance Retrieval:}
Current OCR or STS models are limited by their closed-set training that cannot handle signage recognition with arbitrary shapes, styles, and multi-view inconsistency issues in the open world (see Fig. \ref{fig:dataset}). To address this, we leverage the text set of the venue maps to perform visual similarity matching that enables the closed-set detectors to adapt to the open world without fine-tuning. Specifically, we employ a two-stage process for signage recognition. In the offline stage, we utilize an off-the-shelf multilingual text-diffusion model, AnyText \cite{tuo2023anytext}, to convert all the landmark names $\mathcal{T}$ to the text-rendered signage images set, denoted as $\{o^\prime\}$. Then, we adopt ESTextSpotter \cite{huang2023estextspotter}, denoted as $\phi$, to detect text regions and only extract the text features $e$ before the recognition heads from these offline-generated images without using the recognition results. This process creates a prior features pool $\mathcal{D}_e$. During online exploration, we utilize the same spotter $\phi$ to detect the signage from the current image observation $o_t$ and extract its features $e_t$ at timestep $t$. We then calculate the cosine similarity $s_\phi (o_t, o^\prime_j)$ between the detected text features and those of the generative text images to find the most similar pair. In practice, we also enhance the retrieval process by narrowing down the matching scope to a subset of $h$-hop neighboring ($NN$) to the previous recognized landmark $\tau_{t-1}$ on the venue maps based on the robot's real-time location, resulting in better performance and faster speed:
\begin{equation}
    s_\phi (o_t, o^\prime_j) = cos(\phi(o_{t}), \phi(o^\prime_j)), \ j \in NN (\mathcal{G}, \tau_{t-1}, h).
\end{equation} 
As such, the detected signage $o_t$ is recognized and matched to the most similar landmark name $\tau_t$ at the image level, i.e., $\tau_t = \mathcal{T}_{arg\max s_\phi (o_t, o^\prime_j)}$, with a recognition score $s_\phi (o_t, o^\prime_j)$. 

\textbf{2D-to-3D Fusion:}
To mitigate the noisy text recognition from single-view 2D images, we propose adopting a 2D-to-3D instance fusion strategy \cite{lu2023ovir} to enhance recognition robustness using multiple observations. During exploration, the signage map serves as a 3D text bank $\mathcal{B}$, which maintains the merged 3D text instances $R^{3D}_j$ and corresponding features $e_j^{3D}$. At each frame $o_t$, we first extract the 2D text features $e_i^{2D}$ for each detected 2D text regions $\{r_i^{2D},..,r_k^{2D}\}$, then project them to 3D regions $\{r_i^{3D},..,r_k^{3D}\}$ using depth $D_t$, pose $p_t$ and camera intrinsics $C$. Afterwards, for each new 3D text region $r_i^{3D}$ and each existing 3D text instance $R^{3D}_j$ in the bank, if their cosine similarity $cos(e^{2D}_i, e^{3D}_j)$ exceeds a predefined threshold $\gamma_s$ and the 3D distance between their centroid points $d(r_i^{3D}, R_j^{3D})$ is lower than a threshold $\gamma_d$, then we fuse two instances by aggregating centroid points and averaging over the features:
\begin{equation}
    R^{3D}_j := R^{3D}_j \cup r_i^{3D} ~\text{and}~e_j^{3D}:=(ne_j^{3D} + e_i^{2D}) / (n+1).
\end{equation}
We also periodically merge and filter redundant 3D instances in the bank following the above conditions.

\vspace{-0.3cm}
\subsection{Exploration-Exploitation Planning}
\label{sec:exploration}

\textbf{Candidate Next Pose:}
During navigating to the next landmark $g_t$, the candidate next poses $c \in V$ are sampled among the set of unvisited frontiers $\mathcal{F}$ and unvisited viewpoints $\mathcal{V}$. The frontiers $f \in \mathcal{F}$ bias the exploration to unknown regions, while the viewpoints $v \in \mathcal{V}$  allow the robot to approach and face the signage for better recognition.

\begin{figure}[t]
    \centering
    \includegraphics[width=0.8\linewidth,height=0.3\textwidth]{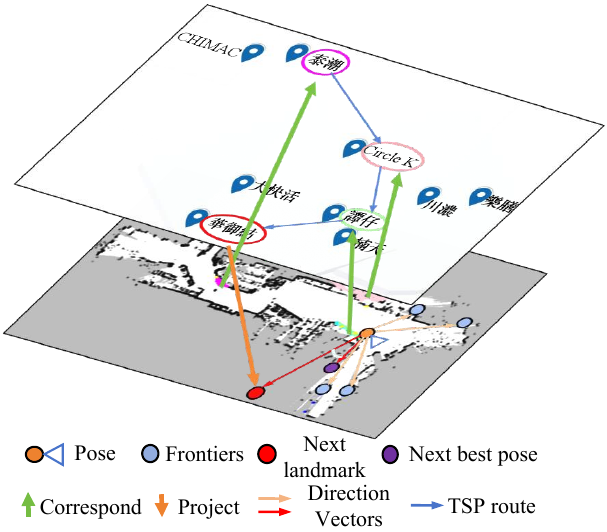}
    \caption{Position estimation of the next landmark by calculating the coordinate transformation between the online map and the venue map for guiding the frontier selection.
    }
    \vspace{-0.4cm}
    \label{fig:heuristic}
\end{figure}

\textbf{Next Best Pose Selection:} We tailor the frontier utility $U_f$ and viewpoint utility $U_v$ for the decision on the candidate selection, respectively. In the initial stage, we induce the robot to explore its surroundings by selecting the frontiers that maximize the information gain $\sum G$ within the camera's FOV. After at least two landmarks are found, we align the online map with the venue map and estimate all the landmark poses $p_{g_t}$ at the world coordinate using random sample consensus (RANSAC) \cite{barath2019magsac} and iterative closest points (ICP).
Therefore, we perform the frontier-based informed search using the relative directions between landmarks on the venue maps. To this end, a directional heuristic is calculated to favor selecting the frontiers closest to the direction towards the subgoal $g_t$ (see Fig. \ref{fig:heuristic}), which is defined as the normalized inner product between the vectors from the current position $p_t$ to the frontier $f_i$ and the vector from the current position $p_t$ to the position of next subgoal $p_{g_t}$. 
Specifically, we first calculate the transformation matrix $T=[R|t]$ and scale $\alpha$, which are updated whenever a new landmark is found. Then, we back-project the landmark positions from the venue map to the world coordinates. As such, the directional heuristic $h(f_i)$ is calculated as $h(f_i) = \frac{(p_t - f_i) \cdot (p_t - T^{-1} p_{g_t})}{\Vert p_t - f_i \Vert \Vert p_t - T^{-1} p_{g_t}\Vert}$, and thereby the frontier utility $U_f(f_i, p_t)$ is computed as:
\begin{equation}
    U_f(f_i, p_t) = \lambda \mu(p_t, f_i) h(f_i) - {\eta} d(p_t,f_i), \ f_i \in \mathcal{F}, 
\end{equation}
where $d(p_t,f_i)$ is the Euclidean distance from the current pose to the candidate pose representing a navigation cost with a weight $\eta$, $\lambda$ is a weight giving more importance to directional heuristic than the navigation cost. 
The hysteresis gain \cite{rrt} $\mu(p_t, f_i)$, which equals $\mu_{gain} > 1$ if frontier points $f_i$ lie within a certain radius $\mu_r$ of current position $p_t$ otherwise equals 1, is adapted to bias selecting frontier points in the surroundings. On the other hand, the viewpoint utility $U_v(v_j, p_t)$ is defined to favor the exploitation in known spaces whenever a text instance with a certain confidence is detected:  
\begin{equation}
    U_v(v_j, p_t) = \beta s_{v}(v_j) - {\eta} d(p_t,v_j), \ v_j \in \mathcal{V},
\end{equation}
where the viewpoint score $s_{v}(v_j)= s_\phi(o_{\tau_j}, o_{\tau_j}^\prime)$, i.e., equals the recognition score of the signage text $\tau_j$ belonging to viewpoint $v_j$ when matched to the venue maps, and $d(p_t,v_j)$ is the navigation cost with a weight $\eta$. A factor $\beta$ that determines to what extent of scores we should highlight the viewpoint candidates than frontiers is employed to handle the exploration-exploitation dilemma. Intuitively, the priority of exploiting signage originates from the fact that it can help localize the robot on the venue maps and facilitate the decision on the next best pose. Finally, we obtain the optimal next pose {$c_{t+1}$} among unvisited frontiers and viewpoints that maximizes the overall utility $U(c_j, p_t)$:
\begin{equation}
U(c_j, p_t)= 
 I_f(c_j) U_f(c_j, p_t) + (1-I_f(c_j)) U_v(c_j, p_t).
\end{equation}
where $c_j \in \mathcal{F} \cup \mathcal{V}$ and $I_f(c_j)=1 \ \text{if} \ c_j \in \mathcal{F} \ \text{else} \ 0$. Then we plan a collision-free path to $c_{t+1}$ using the information RRT* planner from Open Motion
Planning Library~\cite{sucan2012the-open-motion-planning-library} with a fixed replan rate, and follow the route by a local planner that excels in dynamic collision avoidance \footnote{\href{https://www.cmu-exploration.com/}{https://www.cmu-exploration.com/}}. The overall procedure is summarized in Algorithm \ref{algo}.

\section{Experimental Results}
\label{sec:result}

\subsection{Experimental Setup}
We evaluate the proposed system in two large-scale shopping malls with 4 and 9 landmarks, respectively. We implement the system on a Scout-mini mobile robot platform shown in Fig.~\ref{fig:overview}. The sensor suite comprises a Realsense D455 RGB-D camera and an OS0-128 LiDAR. The computing is performed on an Intel NUC mini PC with Intel Core i9-12900H and an NVIDIA Jetson Orin NX with an AI performance of 157 TOPS. The WIFI router builds a local network for communication among the robot platform, sensors, and onboard computers. 

\begin{algorithm}[t]
    \caption{{Signage-aware Exploration using Venue Maps}}
    \label{algo}
    \begin{algorithmic}[1]
        \REQUIRE Venue map $\mathbf{M}$ including landmark names $\mathcal{T}$ and a topological graph $\mathcal{G}$, prior features pool $\mathcal{D}_e$, text spotter $\phi$, RGBD image \{$o_t$, $D_t$\}, LiDAR $\mathcal{X}_t$, signage map $\mathcal{M} = \emptyset$, frontiers $\mathcal{F} = \emptyset$, viewpoints $\mathcal{V} = \emptyset$
        \WHILE{$\mathcal{F} \neq \emptyset \ or\  t = 0$}
        \STATE $o_t, D_t, \mathcal{X}_t \leftarrow$
        getObservation()
        \STATE ${\mathcal{M}}, f_t \leftarrow$ mappingModule($\mathcal{X}_t$) 
        \STATE $\mathcal{F} \leftarrow \mathcal{F} \cup f_t $
        \STATE $R^{2D},e^{2D}, v_t \leftarrow \phi(o_t)$
        \IF{$R^{2D}$ is not None}
            \STATE $R^{3D}, e^{3D} \leftarrow$ 3DFusion($R^{2D}, e^{2D}, D_t, \mathcal{M}$)
            \STATE $\tau_t \leftarrow$ textInstanceRetrieval($e^{3D}, \mathcal{D}_e$)
            \STATE $g_t \leftarrow $ topologicalPlanning($\tau_t, \mathcal{G}$)
            \STATE $\mathcal{V} \leftarrow \mathcal{V} \cup v_t $
            \STATE ${\mathcal{M}} \leftarrow$ semanticProjection($\tau_t, R^{3D}$) 
        \ENDIF
        \STATE $c_{t+1} \leftarrow$ getNextBestPose($\mathcal{F}, \mathcal{V}, g_t$)
        \STATE goTo($c_{t+1}$)
        \STATE $t \leftarrow t + 1$
        \ENDWHILE
        \RETURN a signage map $\mathcal{M}$
    \end{algorithmic}
\end{algorithm}

\vspace{-0.4cm}
\subsection{Implementation Details} 
We obtain the venue maps of the two scenarios via Google My Maps\footnote{\href{https://www.google.com/maps/d/}{https://www.google.com/maps/d/}} 
to showcase the landmarks of interest. We use a TSP solver \cite{hagberg2008exploring} to plan the landmark routes on the venue maps. We adopt FAST-LIO2~\cite{xu2021fastlio2} for robot pose estimation, a dynamic object removal module~\cite{fan2023s2mat} for point cloud segmentation, and OctoMap~\cite{hornung2013octomap} for constructing a 3D volumetric map simultaneously. We then squeeze the 3D map into a 2D occupancy grid map. We detect and filter the frontiers by~\cite{rrt}. We generate the viewpoints by heading toward the signage at a proper distance. We adopt MAGSAC~\cite{barath2019magsac} with Levenberg-Marquardt refinement to estimate the transformation matrix between the venue map and world coordinates. The hyperparameters are listed in Table \ref{hyperparameters}, where $\eta$ is chosen based on the distance range of the frontiers in the scenes; $\lambda$ reduces the bias caused by viewpoints often being closer to the robot than frontiers; $\mu_{radius}$ and $\mu_{gain}$ defines the range of frontiers within the robot's surroundings. 

To alleviate the influence of pedestrians during robot exploration, a dynamic object removal approach inherited from~\cite{fan2023s2mat} is deployed, where the front-end sub-module uses a range-image-based method to coarsely detect the points from dynamic objects, followed by a static mapping framework in the back-end sub-module. Each frame from the LiDAR data stream is effectively separated into dynamic (pedestrians) and static (building structures) scan points, while only the static part is sent into OctoMap~\cite{hornung2013octomap}.

We prompt AnyText~\cite{tuo2023anytext} model to generate the text images by: \textit{a sign of a store with} \textbf{[TEXT]} \textit{written on it}, which means: 
\begin{CJK}{UTF8}{gbsn}
一个店铺标识，写着``\textbf{[TEXT]}''
\end{CJK} in Chinese, where \textbf{[TEXT]} is a place name extracted from the venue map using CnOCR\footnote{\href{https://github.com/breezedeus/cnocr}{https://github.com/breezedeus/cnocr}}. 

\begin{table}[t]

    \begin{center}
    \scalebox{0.85}{
        \begin{tabular}{ccc}
            \toprule
            & Hyperparamter & Value \\
            \midrule
            $h$ & num of hop neighboring to previous landmark & 1 \\
            FOV & horizontal field-of-view of camera & 70.5$^{\circ}$ \\
            $\gamma_{map}$ & threshold for merging topological graph & 150 \\
            $\gamma_s$ & similarity threshold for fusion & 0.5 \\
            $\gamma_d$ & distance threshold for fusion & 3 \\
            $\eta$ & weight for navigation cost & 0.1 \\
            $\lambda$ & weight for directional heuristic & 5\\
            $\mu_{gain}$ & constant in hysteresis gain & 5\\
            $\mu_{radius}$ & radius in hysteresis gain & 10\\
            $\beta$ & balance weight for exploration and exploitation & 9 \\
            $v_{linear}$ & maximum linear speed & 0.8 \\
            $v_{angular}$ & angular speed for turning to active perception & 0.1 \\
            \bottomrule
        \end{tabular}
        }
    \end{center}
    \vspace{-0.2cm}
    \caption{Hyperparamters of the robotic system for real-world experiments}
    \label{hyperparameters}
     \vspace{-0.4cm}
\end{table}

\vspace{-0.3cm}
\subsection{Signage Recognition Performance}
\label{exp:acc}
\begin{figure}[t]
    \centering
    \includegraphics[width=0.9\linewidth]{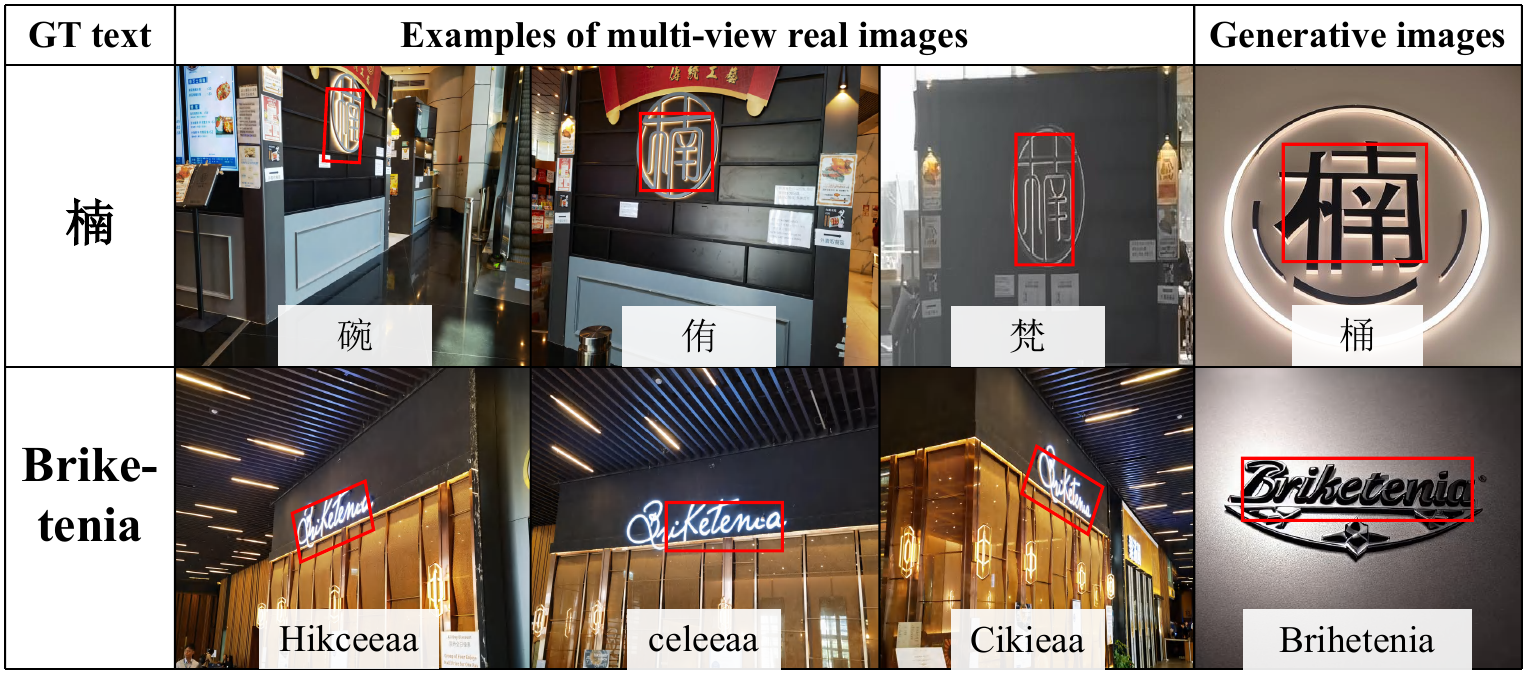} 
    \caption{Examples of signage recognition. The first column shows the ground truths of the texts. In the remaining columns, red boxes highlight the texts of interest, and the words in white boxes are the corresponding recognition results from ESTextSpotter~\cite{huang2023estextspotter}. The noisy recognition results and the interference from text instances that are not presented on the signage make it difficult for accurate signage recognition. Our method can robustly match the real images with generative images even though all their recognized texts are not the same as the ground truths.}
    \label{fig:dataset}
\end{figure}

This experiment is to investigate the signage recognition performance of our proposed signage understanding method. We evaluate our method based on signage recognition recall rates using a dataset of signage images collected in two scenarios, comparing it with four alternative methods: 1) using CLIP~\cite{radford2021learning} to perform text-to-image similarity matching between landmark names and detected images, 2) using Chinese-CLIP \cite{yang2022chinese}, fine-tuned on Chinese dataset, 3) using recognition results of an OCR model, ESTextSpotter \cite{huang2023estextspotter}, with Levenshtein distance as the text-level measurement, and 4) font-based rendering retrieval (FontRR): leveraging the latent features extracted from ESTextSpotter and performing image-level retrieval with font-rendered text images \cite{wen2023visual}. Our method, termed diffusion-based rendering retrieval (DiffusionRR), utilizes a text-diffusion model for text-to-image generation. We collect the signage images of 18 shops and restaurants with 5 different views in real-world shopping malls. Some examples of signage recognition in open-world scenarios, shown in Fig. \ref{fig:dataset}, illustrate the challenges inherent in this task. We consider a sign to be correctly recognized if the top-$k$ ($k=1,2$) retrieval results contain the ground truth and the similarity score is higher than a threshold $\gamma_{recall}=0.3$ for filtering out texts that are neither displayed on signage nor within the text set of venue maps.

\begin{table}[h]
    \centering
    \vspace{-0.1cm}
    \scalebox{0.8}{
    \begin{tabular}{ccc}
        \toprule
        Methods & Recall@1 & Recall@2 \\
        \midrule
        CLIP \cite{radford2021learning} & 28.9\% & 35.6\% \\
        Chinese-CLIP \cite{yang2022chinese} & 28.9\%& 45.6\% \\
        ESTextSpotter \cite{huang2023estextspotter} & 
        52.2\% & 52.2\% \\
        FontRR \cite{wen2023visual} &  62.2\% & 73.3\% \\
        \textbf{DiffusionRR (Ours)} & \textbf{77.8}\% & \textbf{87.8}\% \\
        \bottomrule
    \end{tabular}                                                                           
    }
    \caption{Recall rates of signage recognition ($\uparrow$)}
    \vspace{-0.4cm}
    \label{exp:recall}
\end{table}

The results are reported in Table \ref{exp:recall}. It turns out that the signage recognition capabilities of both CLIP and Chinese-CLIP (ViT-B/16) are poor, achieving only 28.9\% of recall@1. We attribute this to the inability of CLIP models to extract text features from signage images effectively. As a result, they produce similar similarity scores when matching different landmark names, making it difficult to filter out irrelevant texts in the scene. Moreover, the state-of-the-art STS method \cite{huang2023estextspotter} still struggles with signage recognition, achieving only 52.2\% recall rates due to the noisy recognition results. We also show that the most common text measurement, Levenshtein distance, is infeasible for non-alphabet square-shaped characters since it does not consider the similarity of strokes inside the characters. By converting texts to images using font, e.g., Arial, and calculating cosine similarity with signage images \cite{wen2023visual}, the recall rates are improved by 10\%. Furthermore, our diffusion-based approach yields an additional 15.6\% improvement over the font-rending method. We think it is because the generative images are more realistic and have more contextual features, proving that our approach improves the signage recognition capability of a closed-set detector without fine-tuning. Notably, our method's recall@2 is 10\% higher than recall@1, demonstrating better recognition capability when the matching scope is narrowed to neighboring landmarks. By adopting the 2D-to-3D fusion strategy, our method achieves more robust results with higher similarity scores. 

\vspace{-0.4cm}
\subsection{Signage Coverage Efficiency}
\label{exp:mapping}

This experiment is to evaluate the improvement in the efficiency of covering all the landmarks by using venue maps and exploitation method, comparing with Multiple Rapidly Exploring Randomized Trees (RRTs)-based exploration \cite{rrt}, which selects the frontiers by a utility similar to $U_f(f_i, p_t)$ but with substitution of $h(f_i)$ by information gain $\sum G$. We also equip RRTs-based exploration with the proposed signage understanding module. We consider a landmark to be successfully covered if its corresponding signage is accurately recognized, thus here we use signage to represent landmarks. We choose two starting points in each scenario and conduct three trials at each point for two methods, respectively. We evaluate the signage coverage efficiency by both \textit{signage coverage rates} and \textit{exploration time per sign} with the standard variance. The former refers to the average number of recognized signs of all the signs in the scene, and the latter is calculated as $\frac{1}{n}\sum T(i)/S(i)$, where $n$ is the number of trials, $T(i)$ and $S(i)$ are the total exploration time until no detected frontiers and signage coverage number of the $i$-th trial, respectively. We first compare our entire system with the baseline \cite{rrt} and then study the respective impact of venue maps and exploitation in the balance.
\begin{figure}[t]
    \centering
    \includegraphics[width=0.96\linewidth]{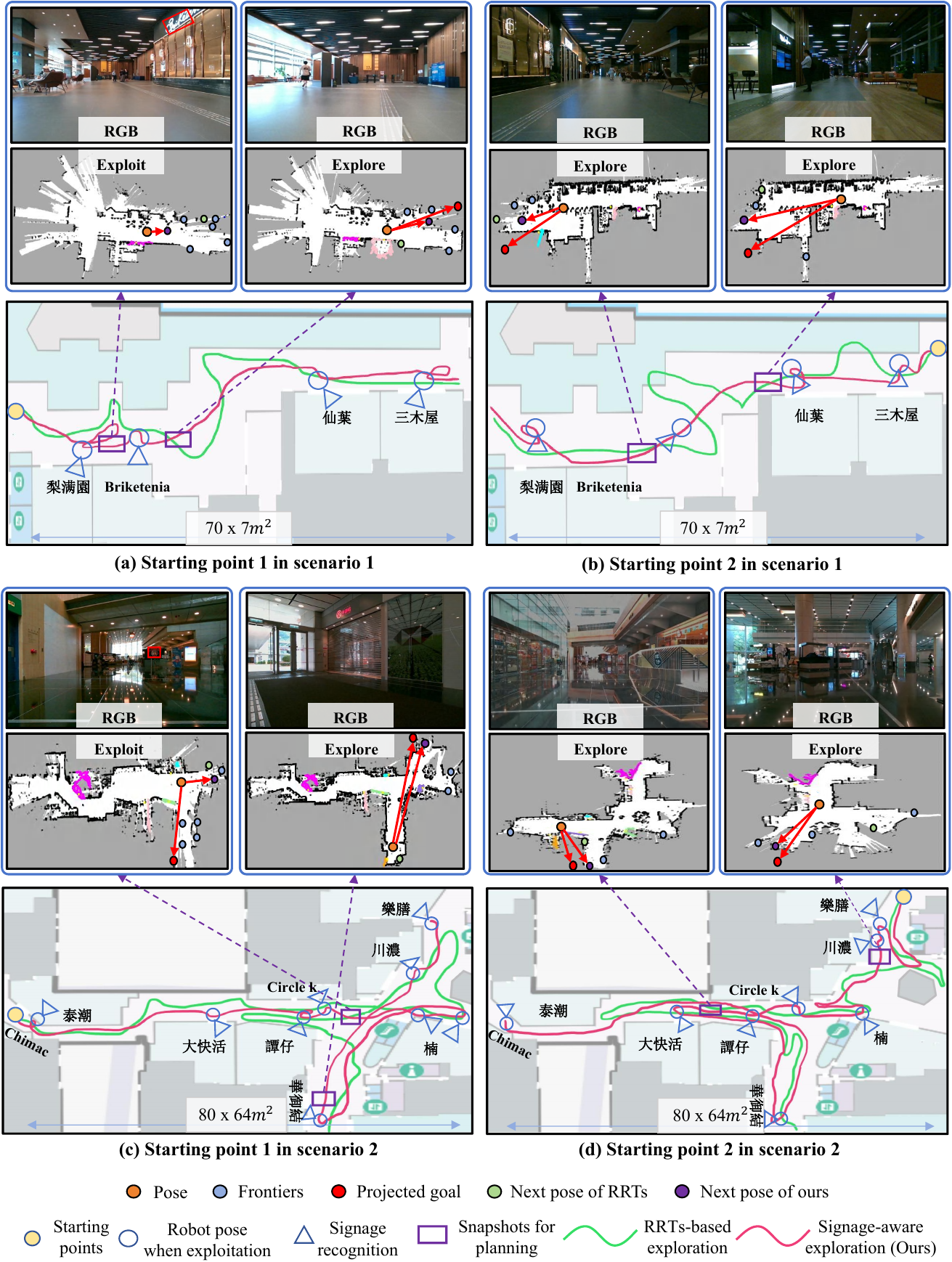}
    \caption{Qualitative examples of the exploration trajectories in four scenario settings. Our method produces more efficient and reasonable exploration paths by using venue maps.}
    \vspace{-0.2cm}
    \label{fig:qualitative}
\end{figure}

\begin{table}[t]
    \centering
    \scalebox{0.6}{
    \begin{tabular}{ccccc}
        \toprule
        
         Scenario 1& \multicolumn{2}{c}{Starting point 1}  &  \multicolumn{2}{c}{Starting point 2}\\  
        (4 landmarks) & Coverage rates $\uparrow$ & Average time ($s$) $\downarrow$ & Coverage rates $\uparrow$ & Average time ($s$)  $\downarrow$ \\
        \midrule
        \cite{rrt} &  1.50~$\pm$~0.71\ /\ 4 & 167.55~$\pm$~25.51 & 1.67~$\pm$~0.58\ /\ 4 & 162.21~$\pm$~70.67 \\
        \textbf{Ours} &  \textbf{3.00~$\pm$~0.00\ /\ 4} &  \textbf{67.89~$\pm$~10.78} & \textbf{3.67~$\pm$~0.58\ /\ 4} &  \textbf{74.94~$\pm$~8.70} \\
        
        \midrule
        Scenario 2 & \multicolumn{2}{c}{Starting point 1}  &  \multicolumn{2}{c}{Starting point 2}\\  
        (9 landmarks)& Coverage rates $\uparrow$ & Average time ($s$) $\downarrow$ & Coverage rates $\uparrow$ & Average time ($s$)  $\downarrow$ \\
        \midrule
        \cite{rrt} &  3.33~$\pm$~0.58\ /\ 9 & 186.67~$\pm$~67.79 & 4.00~$\pm$~1.00\ /\ 9 & 171.55~$\pm$~47.73\\
        
        \textbf{Ours} &  \textbf{7.00~$\pm$~1.00\ /\ 9} &  \textbf{93.40~$\pm$~14.23} & \textbf{6.67~$\pm$~0.58 \ /\ 9} & \textbf{120.58~$\pm$~30.22}\\
        
        \bottomrule
    \end{tabular}
    }
    \vspace{-0.1cm}
     \caption{{Coverage rates and exploration time per sign.}}
    \vspace{-0.4cm}
    \label{exp:semantic coverage}
\end{table}

\begin{figure}[t]
    \centering
    \includegraphics[width=0.49\textwidth, height=0.3\textwidth]{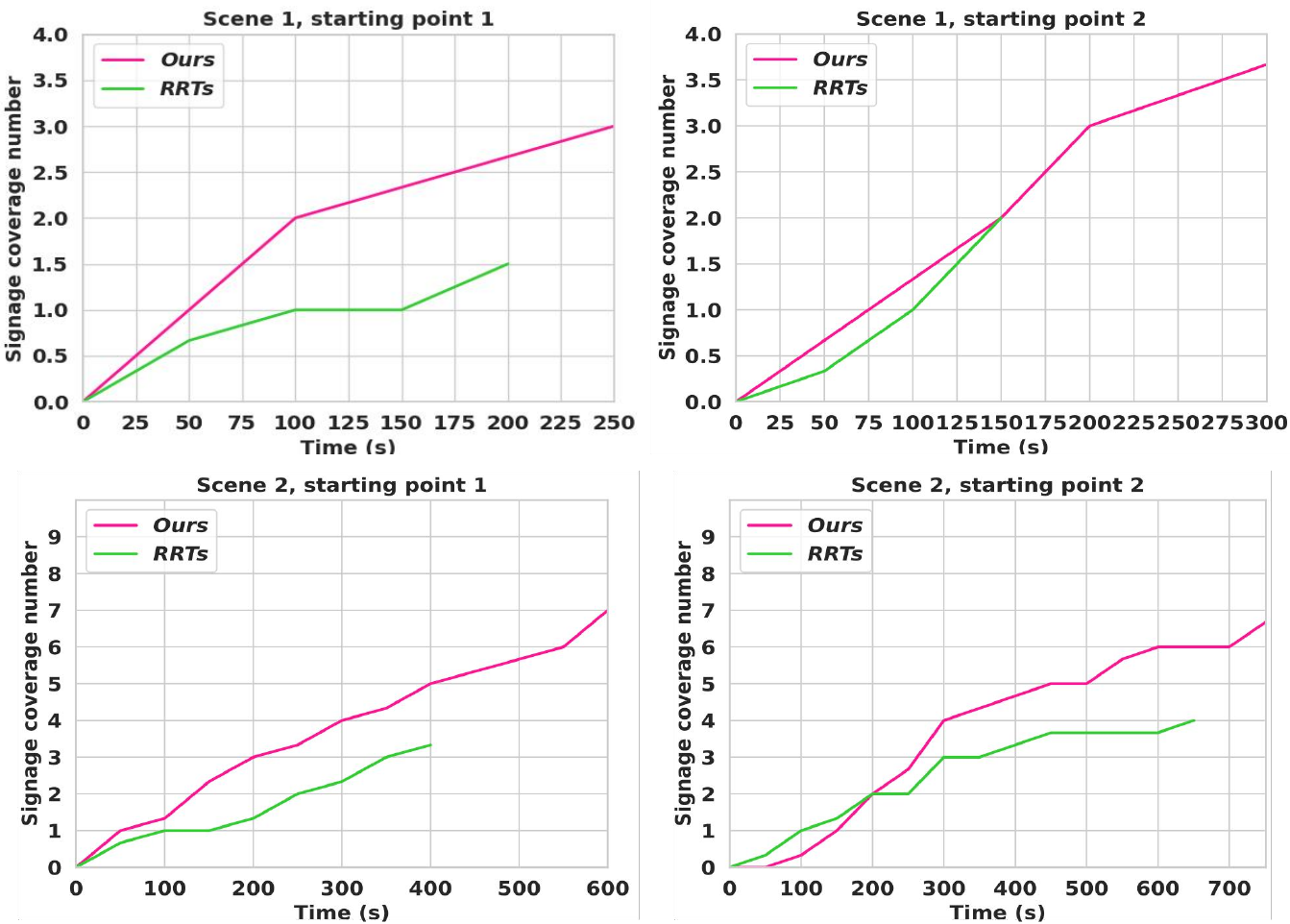}
    \vspace{-0.6cm}
    \caption{The exploration progresses in four scenarios. The curves stop when the last new sign is covered, after which the exploration may continue but no more sign is covered. }
    \vspace{-0.4cm}
    \label{fig:curve}
\end{figure}

The experiment results for comparing our entire system with the baseline are reported in Table \ref{exp:semantic coverage}, and the qualitative examples of the trajectories are illustrated in Fig. \ref{fig:qualitative}. We see that the RRTs-based method covers around 1 and 1.67 of 4 signs at two starting points in scenario 1, respectively. For the larger scenario 2, it can only find 3 and 4 of 9 signs. This is because the RRTs-based method can only recognize the signs when the robot faces them with proper orientations while missing recognizing some signs. Our method successfully covers more signs (3 and 3.67 of 4 signs in scenario 1 and 7 and 6.67 of 9 signs in scenario 2) thanks to the exploitation behavior, enabling the robot to approach the signs and adjust the orientation for better recognition. On the other hand, in Fig. \ref{fig:qualitative}, the RRTs-based method shown in green trajectories blindly explores unknown regions with many redundant steps, yielding inefficient paths and spending about 168$s$ and 162$s$ in scenario 1 and 187$s$ and 172$s$ in scenario 2, respectively, for searching for each sign. By using venue maps, our method in red trajectories localizes the robots on the venue maps after recognizing the sign, thereby knowing the approximate locations of the next signs by calculating the directional heuristic. Therefore, our method navigates to the signs quickly within the narrowed regions, which results in a reduction of nearly 1x the exploration time per sign compared to the baseline and produces more efficient exploration paths to cover the signs (see Fig. \ref{fig:curve}).

\begin{figure}[t]
    \centering
    \includegraphics[width=0.96\linewidth]{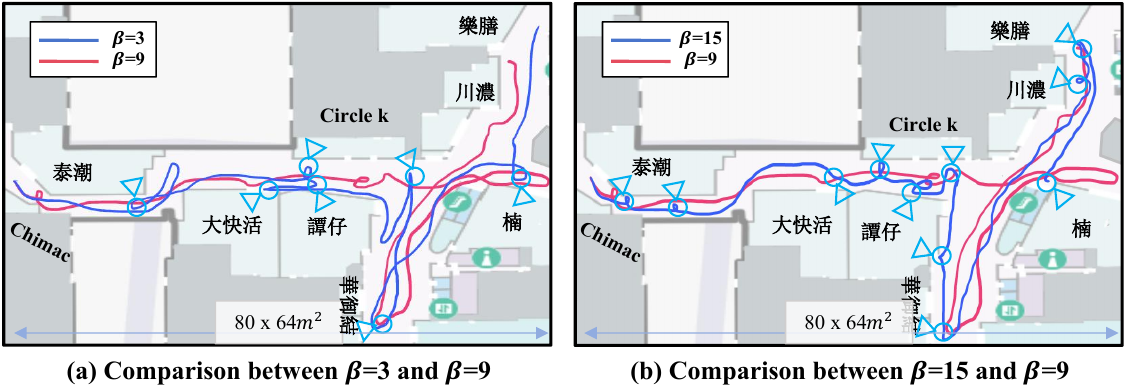}
    \vspace{-0.2cm}
    \caption{Qualitative examples for evaluating the impact of balance weight $\beta$. Triangles and circles show the views and poses when performing exploitation with $\beta=3$ and $15$.}
    \vspace{-0.1cm}
    \label{fig:ablation}
\end{figure}

An important design choice is the weighting parameter $\beta$, which balances venue map-guided exploration and exploitation.  To investigate the impact of venue maps and exploitation, we evaluate $\beta$ with values $\{3, 9, 15\}$. A higher weight assigns greater importance to exploitation. The experimental results tested in scenario 2 starting from point 1 are reported in Table \ref{exp:ablation} and the corresponding trajectories are compared in Fig. \ref{fig:ablation}. When the weight is too low ($\beta=3$), the planner is approximately degraded to the RRT-based exploration with only venue maps. While inclining to frontier exploration, it often fails to recognize certain signs due to poor observations. Although it can eventually revisit part of the missed signs using venue maps, it often requires more time for the robot to localize itself in the venue maps, which significantly decreases the utility of venue maps and the overall path efficiency (see Fig. \ref{fig:ablation}a). On the other hand, as the weight is too high ($\beta=15$), the planner is approximately degraded to the RRT-based exploration with only exploitation that it prioritizes selecting viewpoints for perceiving the potential signage. While this enables the robot to cover more signage and frequently localizes itself in the venue maps, excessive exploitation constrains the robot from exploring beyond the surroundings, hindering discovering the distant signs (see Fig. \ref{fig:ablation}b). A moderate balance of exploration and exploitation ($\beta=9$ by default) can achieve both higher signage coverage rates and search efficiency. Videos of real-world experiments are shown on our project website.

\begin{table}[t]
    \centering
    \scalebox{0.78}{
    \begin{tabular}{cccc}
        \toprule
        &Methods & Coverage rates $\uparrow$ & Average time ($s$) $\downarrow$ \\
        \midrule        
        \multirow{4}{*}{\makecell{Scenario 2\\starting point 1}} &  \cite{rrt} &  3.33~$\pm$~0.58\ /\ 9 & 186.67~$\pm$~67.79 \\
        & Ours ($\beta=3$) &  6.00~$\pm$~1.00 /\ 9&   143.17~$\pm$~20.10\\
        & Ours ($\beta=15$) & 6.67~$\pm$~0.58 /\ 9 & 123.99~$\pm$~14.01\\
        & \textbf{Ours ($\beta=9$)} & \textbf{7.00~$\pm$~1.00\ /\ 9} &  \textbf{93.40~$\pm$~14.23} \\
        \bottomrule
    \end{tabular}               
    }
    
    \caption{Ablation results of balance weight $\beta$. Higher weight biases more exploitation behaviors than exploration.}
    \label{exp:ablation}
    \vspace{-0.4cm}
\end{table}

\vspace{-0.2cm}
\section{Discussion}
\textbf{1) Computational overhead.}
Our system relies on lightweight algorithms optimized for real-time applications. The text detection and retrieval processes run asynchronously, consuming approximately 5GB of GPU memory and achieving a speed of $\sim$2fps on our edge device (157 TOPS). This ensures high performance in signage understanding with minimal delay. \textbf{2) Scalability.} Our approach may be less effective in large scenes with sparse signage, as the lack of signs can hinder localization. While our approach emphasizes the effectiveness of leveraging textual information in the scene, one can always integrate our approach into conventional ones to leverage the spatial structures for global localization. In scenes with dense signage, while false positives may increase, our system includes a RANSAC step to eliminate spatially distant matches. Additionally, dense signage can benefit text-absent issues, aiding in initial localization and exploration. \textbf{3) Multilingualism.} In our scenes, some signs display texts in multiple languages, e.g., Japanese and Korean, which can interfere with accurately recognizing target texts in Chinese and English. Our method addresses this through appearance-based similarity matching, even though our text detector has not been trained on these additional languages. 
Fine-tuning the text detectors on multilingual datasets can further enhance multilingual recognition, as AnyText \cite{tuo2023anytext} supports rendering multilingual text on generative images. \textbf{4) Limitations.} Our system may fail to detect signs due to camera over-exposure. This is particularly relevant for signs made of LED lights, where careful adjustment of the exposure rate is crucial. Incorporating shopfront detection methods \cite{sharifi2020detecting} can also help locate the shops when the signage is significantly obscured.

\section{Conclusion}
\label{sec:conclusion}
We present the first signage-aware exploration method that leverages signage in the scenes and the 2D non-metric venue maps to incorporate text-level information for searching for landmarks in unknown open-world environments. To overcome the challenge of signage recognition, we proposed a diffusion-based text instance retrieval method that detects and recognizes signage with arbitrary shapes and styles in the scene effectively. A 2D-to-3D semantic fusion strategy is employed to enhance recognition performance. Furthermore, we design a venue map-guided exploration-exploitation planner to achieve both high signage coverage rates and search efficiency. Real-world experiments demonstrate our method is more efficient and robust compared to the baselines.
\vspace{-0.2cm}

\bibliographystyle{IEEEtran}
\bibliography{main}

\begin{thebibliography}{10}
\providecommand{\url}[1]{#1}
\csname url@samestyle\endcsname
\providecommand{\newblock}{\relax}
\providecommand{\bibinfo}[2]{#2}
\providecommand{\BIBentrySTDinterwordspacing}{\spaceskip=0pt\relax}
\providecommand{\BIBentryALTinterwordstretchfactor}{4}
\providecommand{\BIBentryALTinterwordspacing}{\spaceskip=\fontdimen2\font plus
\BIBentryALTinterwordstretchfactor\fontdimen3\font minus \fontdimen4\font\relax}
\providecommand{\BIBforeignlanguage}[2]{{%
\expandafter\ifx\csname l@#1\endcsname\relax
\typeout{** WARNING: IEEEtran.bst: No hyphenation pattern has been}%
\typeout{** loaded for the language `#1'. Using the pattern for}%
\typeout{** the default language instead.}%
\else
\language=\csname l@#1\endcsname
\fi
#2}}
\providecommand{\BIBdecl}{\relax}
\BIBdecl

\bibitem{wang2015lost}
S.~Wang, S.~Fidler, and R.~Urtasun, ``Lost shopping! monocular localization in large indoor spaces,'' in \emph{Proceedings of the IEEE International Conference on Computer Vision}, 2015, pp. 2695--2703.

\bibitem{radwan2016you}
N.~Radwan, G.~D. Tipaldi, L.~Spinello, and W.~Burgard, ``Do you see the bakery? leveraging geo-referenced texts for global localization in public maps,'' in \emph{2016 IEEE international conference on robotics and automation (ICRA)}.\hskip 1em plus 0.5em minus 0.4em\relax IEEE, 2016, pp. 4837--4842.

\bibitem{howard2021lalaloc}
H.~Howard-Jenkins, J.-R. Ruiz-Sarmiento, and V.~A. Prisacariu, ``Lalaloc: Latent layout localisation in dynamic, unvisited environments,'' in \emph{Proceedings of the IEEE/CVF International Conference on Computer Vision}, 2021, pp. 10\,107--10\,116.

\bibitem{sarlin2023orienternet}
P.-E. Sarlin \emph{et~al.}, ``Orienternet: Visual localization in 2d public maps with neural matching,'' 2023.

\bibitem{sarlin2024snap}
P.-E. Sarlin, E.~Trulls, M.~Pollefeys, J.~Hosang, and S.~Lynen, ``Snap: Self-supervised neural maps for visual positioning and semantic understanding,'' \emph{Advances in Neural Information Processing Systems}, vol.~36, 2024.

\bibitem{shah2022viking}
D.~Shah and S.~Levine, ``Viking: Vision-based kilometer-scale navigation with geographic hints,'' \emph{arXiv preprint arXiv:2202.11271}, 2022.

\bibitem{fan2023s2mat}
T.~Fan \emph{et~al.}, ``S$^2$mat: Simultaneous and self-reinforced mapping and tracking in dynamic urban scenariosorcing framework for simultaneous mapping and tracking in unbounded urban environments,'' 2023.

\bibitem{rrt}
H.~Umari and S.~Mukhopadhyay, ``Autonomous robotic exploration based on multiple rapidly-exploring randomized trees,'' in \emph{2017 IEEE/RSJ International Conference on Intelligent Robots and Systems (IROS)}, Sept 2017, pp. 1396--1402.

\bibitem{bircher2016receding}
A.~Bircher, M.~Kamel, K.~Alexis, H.~Oleynikova, and R.~Siegwart, ``Receding horizon" next-best-view" planner for 3d exploration,'' in \emph{2016 IEEE international conference on robotics and automation (ICRA)}.\hskip 1em plus 0.5em minus 0.4em\relax IEEE, 2016, pp. 1462--1468.

\bibitem{dang2018autonomous}
T.~Dang, C.~Papachristos, and K.~Alexis, ``Autonomous exploration and simultaneous object search using aerial robots,'' in \emph{2018 IEEE Aerospace Conference}.\hskip 1em plus 0.5em minus 0.4em\relax IEEE, 2018, pp. 1--7.

\bibitem{chaplot2020object}
D.~S. Chaplot, D.~P. Gandhi, A.~Gupta, and R.~R. Salakhutdinov, ``Object goal navigation using goal-oriented semantic exploration,'' \emph{Advances in Neural Information Processing Systems}, vol.~33, pp. 4247--4258, 2020.

\bibitem{papatheodorou2023finding}
S.~Papatheodorou, N.~Funk, D.~Tzoumanikas, C.~Choi, B.~Xu, and S.~Leutenegger, ``Finding things in the unknown: Semantic object-centric exploration with an mav,'' in \emph{2023 IEEE International Conference on Robotics and Automation (ICRA)}.\hskip 1em plus 0.5em minus 0.4em\relax IEEE, 2023, pp. 3339--3345.

\bibitem{lu2024semantics}
L.~Lu, Y.~Zhang, P.~Zhou, J.~Qi, Y.~Pan, C.~Fu, and J.~Pan, ``Semantics-aware receding horizon planner for object-centric active mapping,'' \emph{IEEE Robotics and Automation Letters}, 2024.

\bibitem{gu2023conceptgraphs}
Q.~Gu \emph{et~al.}, ``Conceptgraphs: Open-vocabulary 3d scene graphs for perception and planning,'' 2023.

\bibitem{werby2024hierarchical}
A.~Werby, C.~Huang, M.~B{\"u}chner, A.~Valada, and W.~Burgard, ``Hierarchical open-vocabulary 3d scene graphs for language-grounded robot navigation,'' in \emph{First Workshop on Vision-Language Models for Navigation and Manipulation at ICRA 2024}, 2024.

\bibitem{maye2010inferring}
J.~Maye, L.~Spinello, R.~Triebel, and R.~Siegwart, ``Inferring the semantics of direction signs in public places,'' in \emph{2010 IEEE International Conference on Robotics and Automation}.\hskip 1em plus 0.5em minus 0.4em\relax IEEE, 2010, pp. 1887--1892.

\bibitem{shaikh2013ocr}
S.~N. Shaikh and N.~V. Londhe, ``Ocr based mapless navigation method of robot,'' \emph{International Journal of Inventive Engineering and Sciences}, vol.~1, no.~8, pp. 6--12, 2013.

\bibitem{talbot2020robot}
B.~Talbot, F.~Dayoub, P.~Corke, and G.~Wyeth, ``Robot navigation in unseen spaces using an abstract map,'' \emph{IEEE Transactions on Cognitive and Developmental Systems}, vol.~13, no.~4, pp. 791--805, 2020.

\bibitem{case2011autonomous}
C.~Case, B.~Suresh, A.~Coates, and A.~Y. Ng, ``Autonomous sign reading for semantic mapping,'' in \emph{2011 IEEE international Conference on Robotics and Automation}.\hskip 1em plus 0.5em minus 0.4em\relax IEEE, 2011, pp. 3297--3303.

\bibitem{mantelli2021semantic}
M.~Mantelli \emph{et~al.}, ``Semantic active visual search system based on text information for large and unknown environments,'' \emph{Journal of intelligent \& robotic systems}, vol. 101, no.~2, p.~32, 2021.

\bibitem{wang2015bridging}
H.-C. Wang, C.~Finn, L.~Paull, M.~Kaess, R.~Rosenholtz, S.~Teller, and J.~Leonard, ``Bridging text spotting and slam with junction features,'' in \emph{2015 IEEE/RSJ International Conference on Intelligent Robots and Systems (IROS)}.\hskip 1em plus 0.5em minus 0.4em\relax IEEE, 2015, pp. 3701--3708.

\bibitem{hong2019textplace}
Z.~Hong, Y.~Petillot, D.~Lane, Y.~Miao, and S.~Wang, ``Textplace: Visual place recognition and topological localization through reading scene texts,'' in \emph{Proceedings of the IEEE/CVF International Conference on Computer Vision}, 2019, pp. 2861--2870.

\bibitem{tang2022image}
X.~Tang \emph{et~al.}, ``Image retrieval for visual localization via scene text detection and logo filtering,'' in \emph{2022 7th International Conference on Image, Vision and Computing (ICIVC)}.\hskip 1em plus 0.5em minus 0.4em\relax IEEE, 2022, pp. 662--668.

\bibitem{zimmerman2022robust}
N.~Zimmerman, L.~Wiesmann, T.~Guadagnino, T.~L{\"a}be, J.~Behley, and C.~Stachniss, ``Robust onboard localization in changing environments exploiting text spotting,'' in \emph{2022 IEEE/RSJ International Conference on Intelligent Robots and Systems (IROS)}.\hskip 1em plus 0.5em minus 0.4em\relax IEEE, 2022, pp. 917--924.

\bibitem{li2023textslam}
B.~Li, D.~Zou, Y.~Huang, X.~Niu, L.~Pei, and W.~Yu, ``Textslam: Visual slam with semantic planar text features,'' \emph{IEEE Transactions on Pattern Analysis and Machine Intelligence}, 2023.

\bibitem{kirillov2023segment}
A.~Kirillov \emph{et~al.}, ``Segment anything,'' 2023.

\bibitem{radford2021learning}
A.~Radford \emph{et~al.}, ``Learning transferable visual models from natural language supervision,'' in \emph{International conference on machine learning}.\hskip 1em plus 0.5em minus 0.4em\relax PMLR, 2021, pp. 8748--8763.

\bibitem{huang2023estextspotter}
M.~Huang \emph{et~al.}, ``Estextspotter: Towards better scene text spotting with explicit synergy in transformer,'' in \emph{Proceedings of the IEEE/CVF International Conference on Computer Vision}, 2023, pp. 19\,495--19\,505.

\bibitem{li2022ppocrv3}
C.~Li \emph{et~al.}, ``Pp-ocrv3: More attempts for the improvement of ultra lightweight ocr system,'' 2022.

\bibitem{zhang2020adaptive}
C.~Zhang, A.~Gupta, and A.~Zisserman, ``Adaptive text recognition through visual matching,'' in \emph{Computer Vision--ECCV 2020: 16th European Conference, Glasgow, UK, August 23--28, 2020, Proceedings, Part XVI 16}.\hskip 1em plus 0.5em minus 0.4em\relax Springer, 2020, pp. 51--67.

\bibitem{wang2021scene}
H.~Wang \emph{et~al.}, ``Scene text retrieval via joint text detection and similarity learning,'' in \emph{Proceedings of the IEEE/CVF Conference on Computer Vision and Pattern Recognition}, 2021, pp. 4558--4567.

\bibitem{wen2023visual}
L.~Wen \emph{et~al.}, ``Visual matching is enough for scene text retrieval,'' in \emph{Proceedings of the Sixteenth ACM International Conference on Web Search and Data Mining}, 2023, pp. 447--455.

\bibitem{chen2024scale}
C.~Chen, Y.~Liu, Y.~Zhuang, S.~Mao, K.~Xue, and S.~Zhou, ``Scale: Self-correcting visual navigation for mobile robots via anti-novelty estimation,'' in \emph{2024 IEEE International Conference on Robotics and Automation (ICRA)}, 2024, pp. 16\,360--16\,366.

\bibitem{tuo2023anytext}
Y.~Tuo \emph{et~al.}, ``Anytext: Multilingual visual text generation and editing,'' \emph{arXiv preprint arXiv:2311.03054}, 2023.

\bibitem{lu2023ovir}
S.~Lu, H.~Chang, E.~P. Jing, A.~Boularias, and K.~Bekris, ``Ovir-3d: Open-vocabulary 3d instance retrieval without training on 3d data,'' in \emph{7th Annual Conference on Robot Learning}, 2023.

\bibitem{barath2019magsac}
D.~Barath, J.~Matas, and J.~Noskova, ``Magsac: marginalizing sample consensus,'' in \emph{Proceedings of the IEEE/CVF conference on computer vision and pattern recognition}, 2019, pp. 10\,197--10\,205.

\bibitem{sucan2012the-open-motion-planning-library}
I.~A. {\c{S}}ucan, M.~Moll, and L.~E. Kavraki, ``The {O}pen {M}otion {P}lanning {L}ibrary,'' \emph{{IEEE} Robotics \& Automation Magazine}, vol.~19, no.~4, pp. 72--82, December 2012, \url{https://ompl.kavrakilab.org}.

\bibitem{hagberg2008exploring}
A.~Hagberg, P.~Swart, and D.~S~Chult, ``Exploring network structure, dynamics, and function using networkx,'' Los Alamos National Lab.(LANL), Los Alamos, NM (United States), Tech. Rep., 2008.

\bibitem{xu2021fastlio2}
W.~Xu, Y.~Cai, D.~He, J.~Lin, and F.~Zhang, ``Fast-lio2: Fast direct lidar-inertial odometry,'' 2021.

\bibitem{hornung2013octomap}
A.~Hornung, K.~M. Wurm, M.~Bennewitz, C.~Stachniss, and W.~Burgard, ``Octomap: An efficient probabilistic 3d mapping framework based on octrees,'' \emph{Autonomous robots}, vol.~34, pp. 189--206, 2013.

\bibitem{yang2022chinese}
A.~Yang, J.~Pan, J.~Lin, R.~Men, Y.~Zhang, J.~Zhou, and C.~Zhou, ``Chinese clip: Contrastive vision-language pretraining in chinese,'' \emph{arXiv preprint arXiv:2211.01335}, 2022.

\bibitem{sharifi2020detecting}
S.~Sharifi~Noorian, S.~Qiu, A.~Psyllidis, A.~Bozzon, and G.-J. Houben, ``Detecting, classifying, and mapping retail storefronts using street-level imagery,'' in \emph{Proceedings of the 2020 international conference on multimedia retrieval}, 2020, pp. 495--501.

\end{thebibliography}

\end{document}